\documentclass[runningheads]{llncs}
\usepackage[fontsize=12pt]{fontsize}
\usepackage{geometry}
\usepackage{graphicx}
\usepackage{amsmath,amssymb} 
\usepackage{color}
\usepackage[figuresright]{rotating}
\usepackage[noend]{algpseudocode}
\usepackage{algorithmicx,algorithm}
\usepackage{tikz}
\usepackage{pgfplots}
\usepackage{comment}
\usepackage{axessibility}  
\usepackage{multirow}
\usepackage{subfigure}
\usepackage{makecell}
\usepackage{pgfplots}
\usetikzlibrary{intersections}
\makeatletter
\DeclareRobustCommand\onedot{\futurelet\@let@token\@onedot}
\def\@onedot{\ifx\@let@token.\else.\null\fi\xspace}
 
\def\ie{\emph{i.e}\onedot} 
 
\def\etc{\emph{etc}\onedot} 
 
\def\etal{\emph{et al}\onedot}
\usepackage{xspace}
\newcommand{\name}[0]{GP\xspace}
\usepackage{xcolor}

\begin{document}
\pagestyle{plain}
\mainmatter
\def\ACCV22SubNumber{717}  
\title{Generating Pseudo-labels Adaptively for Few-shot Model-Agnostic Meta-Learning} 
\author{Guodong Liu\thanks{The first two authors contribute equally.}, Tongling Wang, Shuoxi Zhang, Kun He\thanks{corresponding author.}}
\institute{School of Computer Science and Technology, Huazhong University of Science and Technology\\
\email{\{guodl, wangtl, zhangshuoxi, brooklet60\}@hust.edu.cn}
}
\maketitle

\begin{abstract}
Model-Agnostic Meta-Learning (MAML) is a famous few-shot learning method that has inspired many follow-up efforts, such as ANIL and BOIL. However, as an inductive method, MAML is unable to fully utilize the information of query set, limiting its potential of gaining higher generality. 
To address this issue, we propose a simple yet effective method that  generates psuedo-labels adaptively and could boost the performance of the MAML family. 
The proposed methods, dubbed Generative Pseudo-label based MAML (GP-MAML), GP-ANIL and GP-BOIL, leverage statistics of the query set to improve the performance on new tasks. Specifically, we adaptively add pseudo labels and pick samples from the query set, then re-train the model using the picked query samples together with the support set. 
The GP series can also use information from the pseudo query set to re-train the network during the meta-testing. While some transductive methods, such as Transductive Propagation Network (TPN),  struggle to achieve this goal.
Experiments show that our methods, GP-MAML, GP-ANIL and GP-BOIL, 
can boost the performance of the corresponding model considerably,
and achieve competitive performance as compared to the state-of-the-art baselines.

\end{abstract}

\section{Introduction}
\label{01Introduction}
Labeled classification models have achieved remarkable achievements in various domains, such as classification on images, texts, audio, or videos. Training these classification models often necessitates a large amount of labeled data. However, it is extremely hard or even impossible to obtain the labeled data in some fields, such as medical imaging, military applications, \etc. Such challenge leads to \emph{Few-Shot Learning (FSL)}, that aims to train a model with limited training data in new tasks.

Based on whether the statistic information of the unlabeled query set is considered, FSL can be divided into inductive methods~\cite{MAML,IMAML,ANIL,BOIL,Reptile} and transductive methods~\cite{RelationNetwork,MAML++,TPN}.
From the perspective of the solution approach, FSL models roughly fall into three categories, \ie, optimization-based \cite{MAML,IMAML,ANIL,BOIL,Reptile}, metric-based \cite{MatchingNetwork,PrototypicalNetwork,TPN,RelationNetwork,CloserLook}, and model-based \cite{MemoryAugmented}. 
Since the model-based method is not the focus of our work, we will 
mainly discuss the first two categories. 

The most typical optimization-based method, Model-Agnostic Meta-Learning (MAML)~\cite{MAML}, divides the training data into two parts, \ie, support set and query set. 
Under this setting, the training process consists of two stages: inner loop and outer loop. 
The inner loop effectively optimizes the initial parameters for unseen tasks with limited labeled support training data, whereas the outer loop can access the loss for optimization. 
The second category is metric-based, such as Matching Networks~\cite{MatchingNetwork}, Prototypical Networks~\cite{PrototypicalNetwork}, and Transductive Propagation Network~\cite{TPN}, aiming to learn deep embeddings with strong generalization ability.

However, as an inductive method, MAML is unable to fully utilize data from the query set to improve the performance. 
It is also hard for transductive methods to utilize information from the unlabeled query set to re-train the network in a targeted manner during the meta-test.
To tackle these difficulties, we take the pseudo query data into account.
We investigate the impact of pseudo query data on the performance of the MAML family and observe that: (1) the classifier of MAML family is highly sensitive to the pseudo query data; (2) the feature extractor of MAML family is highly adaptable to the pseudo query data; and (3) the imbalance of pseudo query data may have negative impact. 

Motivated by these observations, we study the method of labeling and picking query data as the pseudo labeled data, and propose a new method called the Generative Pseudo-label based MAML (GP-MAML). We generate pseudo-labels by using  label propagation with adaptive picking for MAML and its two typical variants, ANIL~\cite{ANIL} and BOIL~\cite{BOIL}, resulting in three new methods, GP-MAML, GP-ANIL, and GP-BOIL.
Specifically, we use the feature extractor and label propagation to label the query set, addressing the issue that the classifier of inductive models is sensitive to the pseudo query data. 
As the imbalance of pseudo query data may cause negative impact, we select samples from the pseudo query set to balance the sample quantity for each class.
The network is then re-trained using the pseudo query data and the original support data.
As a result, our model can perform targeted parameter updates based on the pseudo query data in both meta-training and meta-test phases.

The main contributions of this work are fivefold. 
\begin{itemize}
    \item
    To our knowledge, this is the first to incorporate label propagation used in transductive methods to generate pseudo-labels for MAML, a typical inductive method for few-shot learning.
    \item
    Our method can be applied in both meta-training and meta-testing phases. Even for meta-testing, we may re-train the network leveraging statistics from the unlabeled query set, while existing transductive methods can not handle.  
    \item
    We propose to use \textit{adaptive picking} to select instances from the pseudo query set to balance the number of samples for each class, leading to higher performance for models that are even sensitive to these pseudo data.
    \item
    We further apply our Generative Pseudo-label method (\name) to two typical variants of MAML, and improve their performance, demonstrating the applicability of our approach.
    \item
    The evaluation on three benchmark datasets 
    shows that our method outperforms all existing MAML-based methods and achieves competitive performance in comparison with the state-of-the-art few-shot learning methods. 
\end{itemize}


\section{Related Work}
\label{02RelatedWork}
This section reviews some popular methods on few-shot learning for related works according to
the above categories.
We first present induction versus transduction, and then introduce optimization-based methods and metric-based methods. 

\textbf{Induction versus Transduction.} 
Induction is a type of reasoning from observed training cases to the general rules, which is then applied to test cases.
In the context of few-shot learning, it is hard for the inductive methods~\cite{MAML,ANIL,BOIL,Reptile} to generalize from a small set of training data and adapt to unseen tasks.
Another approach to achieve better improvements with limited training data, called transduction or transductive inference, is to consider the relationships between instances in the test set to make predictions on them as a whole. 
The earliest transductive method was introduced by Vapnik~\cite{TransductiveMethod}, whose purpose is to reduce the classification loss on the specific test set. 
Previous theoretical and experimental studies have shown that transductive inference performs significantly better than inductive methods, especially for small training sets. 
Based on the consistency assumption~\cite{Cluster} in the semi-supervised learning problem, Zhou \etal~\cite{LabelPropagation} propose a label inference method, which transfers the labels from labeled data to unlabeled instances guided by the weighted graph of a manifold structure. 
Wang \etal~\cite{LNP} propose Linear Neighborhood Propagation to construct approximated Laplacian matrix to avoid the $\sigma$ setting, which is the key to construct the weighted graph.

\textbf{Optimization-based methods.} 
The optimization-based methods aim to quickly learn the parameters to be optimized when the model comes across new tasks.
As the most far-reaching optimization-based method, MAML~\cite{MAML} trains the model's initial parameters so that the model would have a good performance on new tasks after only a small number of gradient updates. 
Due to the expensive second-order gradient computation in MAML, Nichol \etal~\cite{Reptile} propose a first-order method, called Reptile, that reduces the computational complexity and reaches the same accuracy of MAML. 
Antoniou \etal~\cite{MAML++} summarize many problems existing in the MAML and make the  corresponding improvements for these problems.
Flennerhag \etal~\cite{WarpMAML} introduce the expressive capacity and flexibility of memory-based meta-learners into MAML to facilitate gradient descent across the task distribution.
Meta-MinibatchProx is proposed by Zhou \etal~\cite{MinibatchProx} for quickly converging to the optimal hypothesis by learning a prior hypothesis shared across tasks.
ANIL (Almost No Inner Loop)~\cite{ANIL} only updates the classifier parameters of MAML in the inner loop and almost removes the inner loop without reduction in performance. 
Jaehoon \etal~\cite{BOIL} show that the success of the MAML family is attributed to the reuse of high-quality features from the meta-initialized parameters and introduce a simple yet effective algorithm called BOIL (Body Only update in Inner Loop) that only updates the extractor parameters during the inner loop and exhibits significant performance improvement over MAML, especially on cross-domain tasks.

\textbf{Metric-based methods.} 
This stream of methods converts the query image and support image to the same embedding space. 
It then classifies the query images by calculating the distance or similarity among the embedding features.
Vinyals \etal~\cite{MatchingNetwork} train a weighted nearest neighbor classifier by the support set and update the model according to the performance on query set.
Snell \etal~\cite{PrototypicalNetwork} introduce Prototypical Networks, which calculate a prototype for each class in the support set and predict the category of the query set by Euclidean distance to prototype. 
Sung \etal~\cite{RelationNetwork} propose Relation Networks that calculate the distance between query and support samples by learning a deep non-linear distance metric.
Liu \etal~\cite{TPN} propose TPN (Transductive Propagation Network) that propagates labels from labeled support instances to unlabeled query instances by learning a graph construction module that exploits the manifold structure in the data.
TPN alleviates the low-data problem by utilizing the whole query set for prediction.

In contrast with the above approaches, our method provides a transductive way to make fully use of query set data in the inductive method.
To the best of our knowledge, we are the first to incorporate the label propagation to the MAML family and take the pseudo query data into account.
Experiments show our framework outperforms all existing methods based on MAML.

\section{Methodology}
\label{03Method}
In this section, we first introduce in detail the two most related methods, the MAML series and TPN, and then present our proposed GP-MAML.


\subsection{MAML and Its Variants}  
Model-Agnostic Meta-Learning (MAML) tries to learn an effective meta-initialization so that the model can start from this initialization and achieve excellent results with limited data training. 
Unlike traditional machine learning models that are trained directly on labeled data, MAML is trained on many tasks $\mathcal{T}$ sampled from the dataset distribution  $p(\mathcal{T})$. 
Each task $\mathcal{T}_i$ consists of a support set $spt_{i}$ and a query set $qry_i$. The support set samples under the N-way-K-shot setting (each support set consists of $N$ classes and each class has $K$ samples), the query set contains unlabeled data for evaluation.

The training procedure of MAML includes two stages: the inner loop and the outer loop. In the inner loop, the model first computes the task-specific loss $\mathcal{L}_{spt_{i}}\left(f_{\theta}\right)$ on the support set, where $f_{\theta}$ is a neural network parameterized by $\theta$. Then MAML conducts a task-specific gradient update on the model:
\begin{align}
    \theta_i^{\prime}=\theta-\alpha \nabla_{\theta} \mathcal{L}_{spt_{i}}\left(f_{\theta}\right) .
    \label{Equation1}
\end{align}
In the outer loop, MAML first computes the loss $\mathcal{L}_{qry_{i}}\left(f_{\theta_{i}^{\prime}}\right)$ of each query set based on the updated parameter of the inner loop. Then the outer loss of all the tasks in a batch is calculated as $\sum_{\mathcal{T}_{i} \sim p(\mathcal{T})} \mathcal{L}_{qry_{i}}\left(f_{\theta_{i}^{\prime}}\right)$ . 
The meta-initialized parameters are then updated across the sampled tasks using the gradient descent method with respect to parameter $\theta$:
\begin{align}
    \theta \leftarrow \theta-\beta \nabla_{\theta} \sum_{\mathcal{T}_{i} \sim p(\mathcal{T})} \mathcal{L}_{qry_{i}}\left(f_{\theta_{i}^{\prime}}\right) .
\end{align}

There are two main variants based on MAML, \ie, ANIL~\cite{ANIL} and BOIL~\cite{BOIL}. 
Both of them split the model parameters $\theta$ into two parts, $\theta_{body}$ and $\theta_{head}$, representing parameters for feature extractor and classifier, respectively. 
They have the same outer loop as MAML does.
Yet in the inner loop, ANIL only updates the head parameters $\theta_{head}$ (Equation~\ref{Equation3}),
 while BOIL only updates the body parameters $\theta_{body}$ (Equation~\ref{Equation4}).

\begin{equation} 
\theta_{{head}_{i}}^{\prime}=\theta_{head}-\alpha \nabla_{\theta_{head}} \mathcal{L}_{spt_{i}}\left(f_{\theta_{body}, \theta_{head}}\right),\ \theta_{{body}_{i}}^{\prime}=\theta_{body}\text{,}
\label{Equation3}
\end{equation}

\begin{equation}
\theta_{{body}_{i}}^{\prime}=\theta_{body}-\alpha \nabla_{\theta_{body}} \mathcal{L}_{spt_{i}}\left(f_{\theta_{body}, \theta_{head}}\right),\ \theta_{{head}_{i}}^{\prime}=\theta_{head}\text{.}
\label{Equation4}
\end{equation}



\subsection{Transductive Propagation Network}
Transductive Propagation Network (TPN)~\cite{TPN} is a transductive metric-based method in which label propagation~\cite{LabelPropagation} is critical to the success.
TPN will examine the relationship between support data and query data as well as the relationship between query data when predicting on the query data, 
and propagate labels from labeled support data to unlabeled query data, resulting in significantly higher accuracy than that of inductive methods.

TPN consists of four components: (1) feature embedding with a convolutional neural network $f$, (2) graph construction module $g$ that produces example-wise parameters to exploit the manifold structure, (3) label propagation that spreads labels from the support set $spt$ to the query set $qry$, and (4) a loss generation step that computes a cross-entropy loss between propagated labels and ground-truths on the query set to train all framework parameters jointly.

TPN uses the convolutional neural network $f$ to perform feature extraction on the support set and the query set for a task.
The resulting feature maps $f(\mathbf{x}_i), \mathbf{x}_{i} \in {spt \cup qry}$ are used to calculate the Gaussian similarity between the samples:
\begin{align}
    W_{i j}=\exp \left(-\frac{1}{2} d\left(\frac{f\left(\mathbf{x}_i\right)}{\sigma_{i}}, \frac{f\left(\mathbf{x}_{j}\right)}{\sigma_{j}}\right)\right) ,
\end{align}
where $W \in R^{(N \times K+T) \times(N \times K+T)}$ for all instances in $spt \cup qry$, \space $d(\cdot, \cdot)$ is the Euclidean distance, $\sigma_i=g(f(\mathbf{x}_i))$ is an example-wise length-scale parameter calculated by the graph construction module $g$. 

Label propagation is based on the obtained $W_{i j}$. 
Firstly, it applies the normalized graph Laplacians on $W$, as follows:
\begin{align}
    S=D^{-1 / 2} W D^{-1 / 2} ,
\end{align}
where $D$ is a diagonal matrix whose diagonal values are the sums of the corresponding rows of $W$.
Then, TPN defines a label matrix $Y$ with $Y_{ij}=1$ if $\mathbf{x}_i$ is from the support set and labeled as $\mathbf{y}_i=j$.
The solution for the predicted labels is as follows:
\begin{align}
    F^{*}=(I-\alpha S)^{-1} Y,
\end{align}
where $\alpha \in(0,1)$ is a hyper-parameter controlling the amount of propagated information, and $I$ is the identity matrix~\cite{LabelPropagation}.

\begin{figure}[t]
\centering

\includegraphics[width=0.8\textwidth]{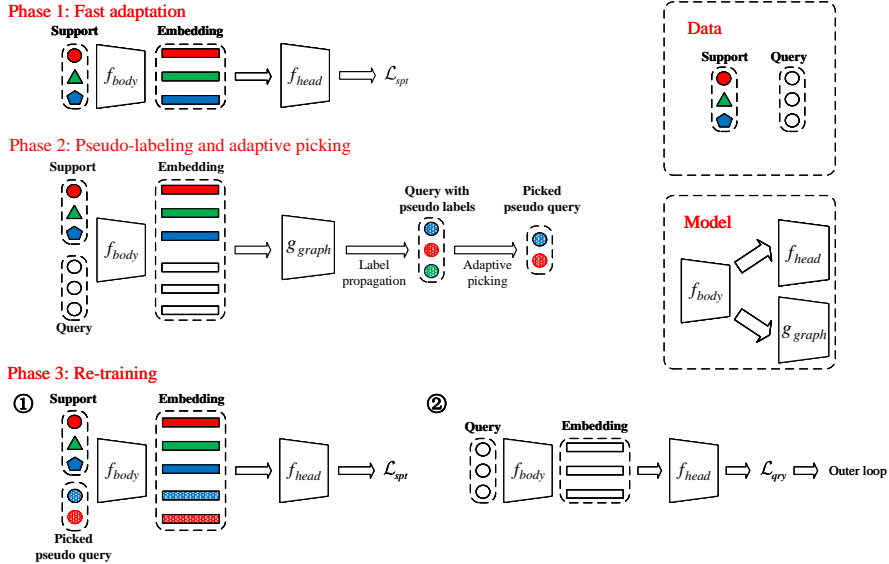}
\caption{The overall framework of GP-MAML. It contains three phases: fast adaptation,  pseudo-labeling and adaptive picking, and re-training. 
In phase 1, the few-shot model updates its parameters through the support set. 
In phase 2, we adopt the label propagation of TPN to label the query set, which will be filtered by adaptive picking to select the pseudo query data.
In phase 3, the picked pseudo query data will be added to the support set for re-training, then the loss of the query set is calculated as in MAML.}
\label{fig:overallFramework}
\end{figure}

\subsection{The Proposed GP-MAML} 
\label{sec:3_3}
As an inductive method, MAML can not fully utilize the data of the query set to achieve better performance.
Even for transductive methods, it is also hard to update their parameters during meta-testing according to the query set. To tackle these issues, we introduce Generative Pseudo-label (\name) into MAML, making it possible to leverage the statistic information of the query set in both meta-training and meta-testing.


Our framework is illustrated in Fig. \ref{fig:overallFramework}.
The classifier $f_{head}$ and the graph construction module $g_{graph}$ share the same feature extractor $f_{body}$. Our training strategy consists of three phases:

\textbf{Phase 1: Fast adaptation.}
We use the feature extractor $f_{body}$ to extract features of the support set and classify these features using $f_{head}$. 
Then we update the parameters of $f_{body}$ using the loss $\mathcal{L}_{spt}$.

\textbf{Phase 2: Pseudo-labeling and adaptive picking.}
We firstly use label propagation to label the query set.
Specifically, we use the updated $f_{body}$ obtained in phase 1 to re-extract features of the support set and the query set. The resulting features are embedded into a new manifold structure using the graph construction module $g_{graph}$, which is then used for label propagation.
We take the label propagation result of confidence measure (CM) as input and label the data in the query set.
However, the pseudo query data can not be directly used to expand the support set because the number of samples in different classes of the pseudo query data varies greatly.
Thus we propose \textit{adaptive picking}, as shown in Algorithm \ref{algorithm:adaptiveSelect}, by taking the CM and pseudo query data as input.
Then we calculate the smallest sample size (say $k$) among all categories and pick the top $k$ samples for each category according to CM.

\textbf{Phase 3: Re-training.}
All the samples chosen in phase 2, together with the original support set, are then used to re-train $f_{body}$ and $f_{head}$. 
The inner loop of GP-MAML is then competed after calculating the loss $\mathcal{L}_{qry}$ of the original query set
through the re-trained $f_{body}$ and $f_{head}$. 
In the outer loop, we sum the loss $\mathcal{L}_{qry}$ of each task and update $f_{body}$ and $f_{head}$ with the original parameters as MAML does. 
Details of GP-MAML are presented in Algorithm \ref{algorithm:ourMethod}.
We also generate pseudo-labels for ANIL and BOIL by using label propagation with adaptive picking, denoted as GP-ANIL and GP-BOIL, respectively. The only difference is the process of gradient update in the inner loop, as shown in Equations~\ref{Equation1}, ~\ref{Equation3}, and ~\ref{Equation4}.

\begin{algorithm}[ht]
\caption{The GP-MAML algorithm} 
\hspace*{0.02in} {\bf Require:}  
$p(\mathcal{T})$: distribution over tasks\\
\hspace*{0.02in} {\bf Require:} 
$\alpha, \beta, \gamma$ : hyperparameters on step size\\
\hspace*{0.02in} {\bf Require:} 
$f$: backbone network model with parameters $\theta_{body}$, $\theta_{head}$\\
\hspace*{0.02in} {\bf Require:} 
$g$: graph construction model with parameter $\theta_{graph}$\\
\begin{algorithmic}[1]
\State Randomly initialize $\theta_{body}$, $\theta_{head}$, $\theta_{graph}$
\While{not done} \do\\
    \State Sample a batch of tasks $\mathcal{T}_{i} \sim p(\mathcal{T})$
    \For{\textbf{all} $\mathcal{T}_{i}$} \do\\
        \State $spt_{i}, qry_{i} \leftarrow \mathcal{T}_{i}$ 
        \State Evaluate $\nabla_{\theta_{body}} \mathcal{L}_{spt_{i}}\left(f_{\theta_{body}, \theta_{head}}\right)$ with respect to support set examples
        \State Compute adapted body parameters with gradient descent: $$\theta_{{body}_{i}}^{TPN}=\theta_{body}-\alpha \nabla_{\theta_{body}} \mathcal{L}_{spt_{i}}\left(f_{\theta_{body}, \theta_{head}}\right)$$
        \State Compute embedded feature maps with support set and query set examples and construct the weighted graph:
        $$Graph_i=g_{\theta_{graph}}(f_{\theta_{{body}_{i}}^{TPN}}(spt_i, qry_i))$$
        \State Obtain CM and pseudo query data of $\mathcal{T}_{i}$ using label propagation: $$CM_i, PLabel_i=Label Propagation(Graph_i)$$
        \State Expand support set with picked pseudo query data:
        $$spt_{i}^{\prime} \leftarrow spt_{i}+AdaptivePicking(CM_i,PLabel_{{query}_i})$$
        
        \State Re-update parameters with new support set examples: 
        $$\theta_{{body}_{i}}^{\prime}, \theta_{{head}_{i}}^{\prime} \leftarrow \theta_{body},\theta_{head}-\beta \nabla_{\theta_{body},\theta_{head}} \mathcal{L}_{spt_{i}^{\prime}}\left(f_{\theta_{body}\theta_{head}}\right)$$
    \EndFor
    \State Update $\theta_{body}, \theta_{head} \leftarrow \theta_{body},\theta_{head}-\beta \nabla_{\theta_{body},\theta_{head}} \sum_{\mathcal{T}_{i} \sim p(\mathcal{T})} \mathcal{L}_{qry_{i}}\left(f_{\theta_{{body}_{i}}^{\prime},\theta_{{head}_{i}}^{\prime}}\right)$
    \State Update $\theta_{graph} \leftarrow \theta_{graph}-\gamma \nabla_{\theta_{graph}}\sum_{\mathcal{T}_{i} \sim p(\mathcal{T})}\mathcal{L}_{qry_{i}}\left(g_{\theta_{graph}}\right)$
\EndWhile
\end{algorithmic}
\label{algorithm:ourMethod}
\end{algorithm}

\begin{algorithm}[ht]
\caption{The Adaptive Picking Method} 
\hspace*{0.02in} {\bf Input:}  
$CM$: confidence measure of query set\\
\hspace*{0.02in} {\bf Input:} 
query set $qry=\left\{\left(\mathbf{x}_{1}, \mathbf{y}_{1}\right), \ldots,\left(\mathbf{x}_{N}, \mathbf{y}_{N}\right)\right\}$, where $\mathbf{y}_{i} \in\{1, \ldots, C\}$ is the pseudo label of $\mathbf{x}_i$\\
\hspace*{0.02in} {\bf Output:} 
the picked pseudo query set $P$
\begin{algorithmic}[1]

\State Calculate the number of picked samples in each category:\\
~~~~~$k = \min_{i=1}^{C} \sum_{j=1}^{N} I(\mathbf{y}_j=i)$
\For{$i$ in $\{1, \dots, C\}$} \do\\
    \State Pick the top $k$ confidence samples in category $i$ to join $P$
\EndFor
\State return $P$
\end{algorithmic}
\label{algorithm:adaptiveSelect}
\end{algorithm}

\section{Experiments}
\label{04Experiments}
In this section, following the line of MAML, ANIL, and BOIL, we provide practical details of the method presented in Section ~\ref{sec:3_3} and examine their effectiveness.
We conduct a comprehensive experimental analysis of our GP-MAML, GP-ANIL, and GP-BOIL, and do ablation studies to verify the effectiveness of our proposed Adaptive Picking (AP) method.
The goal of our experiments is to answer the following questions: 
\begin{itemize}
    \item Q1: Can we improve the MAML family  by using the query set? 
    \item Q2: Is it possible to make the pseudo query data more balanced by picking more carefully?
    \item Q3: Is it possible to make the pseudo query data more balanced by taking the relationship among samples into account? 
\end{itemize}

\subsection{Datasets}
We choose four datasets for experiments.   

\textbf{MiniImageNet.} 
The miniImageNet dataset~\cite{MiniImageNet} is the most popular few-shot learning benchmark.
It is composed of $60,000$ images selected from the ImageNet dataset~
\cite{ImageNet}, with a total of $100$ categories. 
Each category has $600$ images, and the size of each image is $84 \times 84$. We follow the class splits used by Ravi and Larochelle~\cite{MiniImageNet}, which include $64$ classes for training, $16$ classes for validation, and $20$ classes for testing. 

\textbf{CIFAR-FS.} 
The full name of CIFAR-FS dataset is the CIFAR100 Few-Shots dataset, which is derived from the CIFAR100 dataset and contains $100$ categories, each having $600$ images, with a total of $60,000$ images. It is usually divided into training set ($64$ categories), validation set ($16$ categories), and testing set ($20$ categories), and the image size is unified to $32 \times 32$.

\textbf{FC100.} 
The Few-shot CIFAR100 dataset (FC100) is similar to the above CIFAR-FS dataset and also comes from the CIFAR100 dataset. The difference is that FC100 is not divided by class but by superclass. There are a total of $20$ superclasses ($60$ categories), including $12$ superclasses in the training set, $4$ superclasses ($20$ categories) in the validation set, and $4$ superclasses ($20$ categories) in the testing set.

\textbf{CUB.}
The Caltech-UCSD Birds-200-2011 dataset (CUB) contains 11,788 images of 200 bird species with 150 seen classes and 50 disjoint unseen classes. It is usually divided into training set ($100$ categories), validation set ($50$ categories), and testing set ($50$ categories), and the image size is unified to $84 \times 84$.

\subsection{Experimental Setup}
 
The backbone few-shot network $F$ is a four convolutional-block network with $64$ channels and a fully-connected layer, following Jaehoon \etal's setting~\cite{BOIL}. 
The graph construction module is composed of two convolutional blocks and two fully-connected layers, according to the TPN's setting~\cite{TPN}, offering an example-wise scaling parameter.
The hyper-parameter $k$ of $k$-nearest neighbor graph is set to $20$, and the $\alpha$ for label propagation is set to $0.99$, as suggested in TPN. We use the batch size of $64$ for all networks. The model is trained with $30,000$ episodes. The initial learning rate of the optimizer is $1e^{-3}$, and drops by a factor of $10$ after $10,000$ and $20,000$ episodes, respectively.
We evaluate the performance of GP-MAML, GP-ANIL, and GP-BOIL in 5-way-1-shot and 5-way-5-shot settings, with 15 queries in each category of the query set.

\subsection{Results and Discussions}

\subsubsection{Improve the MAML family using the query set.}~
For MAML, ANIL and BOIL, the model performs a fast adaptation through the support set, and the updated model is used to calculate the loss of the query set. However, as inductive methods, they can not fully utilize the data statistics from the query set to achieve better performance.
Hence, we first conjecture that the model's performance would be enhanced if it could take the query data into account during the fast adaption process and make targeted updates for the query data.
Specifically, We first utilize the model's own classifier to pseudo-label the query set data after fast adaptation, and then re-train the model with all the pseudo query data and the support set for meta-testing.
By adding the pseudo query data to the support set for fast adaptation, the model can fully consider the relationship between support data and query data, as well as the relationship between query samples.
Table~\ref{tab:QueryDataGain} shows the experimental results.

\begin{table}[t]
    \caption{Comparison on miniImageNet, CIFAR-FS, and FC100 under the 5-way setting. \textit{w. qry} indicates the accuracy when adding the query set to the support set after pseudo-labeling.
    }
    \begin{center}
    \scalebox{0.800}{
    \begin{tabular}{|l|cc|cc|cc|}
        \hline
	    \multirow{2}*{Method} & \multicolumn{2}{c|}{miniImageNet} & \multicolumn{2}{c|}{CIFAR-FS} & \multicolumn{2}{c}{FC100}\vline \\
	    \cline{2-7}
	    & 1-shot & 5-shot & 1-shot & 5-shot & 1-shot & 5-shot \\
	    \hline
	    MAML & $48.24\pm0.32 $ & $61.52\pm1.95$& $57.57\pm0.16$& $72.31\pm0.05 $& $36.64\pm0.05$& $47.26\pm0.13 $ \\
	    ANIL & $49.61 \pm 0.27$ & $64.90\pm 0.63$& $58.34\pm 0.03$& $72.34\pm 0.01$& $36.38\pm 0.16$& $47.16\pm 0.08$ \\
	    BOIL & $49.82\pm  0.24$ & $67.60\pm  0.01$& $58.69\pm 0.11$& $75.31\pm  0.02$& $38.80\pm 0.03$& $51.42\pm 0.20 $ \\
	    \hline
	    MAML (\textit{w. qry}) & $34.89\pm0.41$ & $50.11\pm0.14$& $43.71\pm0.30$& $58.40\pm 0.43$& $25.24 \pm0.15$& $30.51\pm 0.28$ \\
	    ANIL (\textit{w. qry}) & $37.49\pm 0.58$ & $48.41\pm 1.78$& $46.23\pm 0.25$& $59.52\pm 0.08$& $25.68 \pm0.12$& $30.97\pm 0.29$ \\
	    BOIL (\textit{w. qry}) & $\mathbf{49.88\pm 0.46}$ & $\mathbf{68.08\pm 0.15}$& $\mathbf{60.91\pm 0.13}$& $\mathbf{76.32\pm 0.03}$& $\mathbf{39.67\pm 0.02}$& $\mathbf{52.02\pm 0.21}$ \\
	    \hline
    \end{tabular}
    }
    \end{center}
    \label{tab:QueryDataGain}
\end{table}

The results show that after adding the pseudo query data into the support set for fast adaptation, only BOIL maintains its performance, but MAML and ANIL perform poorly. 
The reason for this phenomenon would be related to the \textit{representation reuse} and \textit{representation change}~\cite{BOIL}.
The meta-initialization of MAML and ANIL offers efficient \textit{representation reuse} through the body before fast adaptation, and the body almost does not update its parameters during the fast adaptation.
Despite the fact that the meta-initialization of BOIL provides less efficient \textit{representation reuse} compared to MAML and ANIL, the BOIL's body can update its parameters to extract more efficient representations through task-specific adaptation, which is called the \textit{representation change}.  
This means BOIL is less sensitive to pseudo query data than MAML and ANIL because of the \textit{representation change}.

\subsubsection{Make pseudo query data more balanced by a careful picking.}~
During the experiments, we also observe that the number of samples in different pseudo query data categories varies greatly. So models that are very sensitive to the pseudo query data like MAML and ANIL can not directly use the pseudo query data. To tackle this issue, we propose an adaptive picking method to select as much pseudo query data as possible while maintaining a balanced number of samples for each category.
The comparison results of using adaptive picking are shown in Table~\ref{tab:AdaptivePicking}.

\begin{table}
    \caption{Comparison results on miniImagenetNet, CIFAR-FS, FC100 under the 5-way setting. \textit{w. AP} indicates the accuracy after adaptive picking be used to select pseudo query data to add to the support set.
    }
    \begin{center}
    \scalebox{0.85}{
    \begin{tabular}{|l|cc|cc|cc|}
        \hline
	    \multirow{2}*{Method} & \multicolumn{2}{c|}{miniImageNet} & \multicolumn{2}{c|}{CIFAR-FS} & \multicolumn{2}{c}{FC100}\vline \\
	    \cline{2-7}
	    & 1-shot & 5-shot & 1-shot & 5-shot & 1-shot & 5-shot \\
	    \hline
	    MAML & $48.24\pm0.32 $ & $61.52\pm1.95$& $57.57\pm0.16$& $72.31\pm0.05 $& $36.64\pm0.05$& $47.26\pm0.13 $ \\
	    ANIL & $49.61 \pm 0.27$ & $64.90\pm 0.63$& $58.34\pm 0.03$& $72.34\pm 0.01$& $36.38\pm 0.16$& $47.16\pm 0.08$ \\
	    BOIL & $49.82\pm  0.24$ & $67.60\pm  0.01$& $58.69\pm 0.11$& $75.31\pm  0.02$& $38.80\pm 0.03$& $51.42\pm 0.20 $ \\
	    \hline
	    MAML(\textit{w. AP}) & $51.37\pm0.41$ & $63.82\pm2.39$& $62.58\pm0.27$& $75.08\pm0.05$& $38.65\pm0.02$& $48.92\pm0.12$ \\
	    ANIL(\textit{w. AP}) & $\mathbf{53.00\pm0.33}$ & $67.86\pm0.71$& $63.44\pm0.07$& $75.12\pm0.03$& $38.37\pm0.20$& $48.71\pm0.12$ \\
	    BOIL(\textit{w. AP}) & $52.42\pm0.29$ & $\mathbf{69.84\pm0.01}$& $\mathbf{63.51  \pm0.11}$& $\mathbf{77.76\pm0.09}$& $\mathbf{40.76\pm0.01}$& $\mathbf{52.97\pm0.24}$ \\
	    \hline
    \end{tabular}
    }
    \end{center}
    \label{tab:AdaptivePicking}
\end{table}

\begin{sidewaystable}

    \caption{Comparison on miniImageNet, CIFAR-FS, FC100 and CUB under the 5-way setting. GP-MAML, GP-ANIL and GP-BOIL outperform 
    MAML, ANIL and BOIL, respectively. All accuracy results are averaged over 600 test episodes.
    \textit{BN} means that information is shared between the test samples via batch normalization~\cite{BatchNormalization}.
    }
    \begin{center}
    \scalebox{0.85}{
    \begin{tabular}{|l|cc|cc|cc|cc|}
        \hline
	    \multirow{2}*{Method} & \multicolumn{2}{c|}{miniImageNet} & \multicolumn{2}{c|}{CIFAR-FS} &
	    \multicolumn{2}{c|}{FC100} &
	    \multicolumn{2}{c}{CUB} \vline \\
	    \cline{2-9}
	    & 1-shot & 5-shot  & 1-shot & 5-shot & 1-shot & 5-shot & 1-shot & 5-shot \\
	    \hline
	    MAML~\cite{MAML} & $48.24\pm0.32 $ & $61.52\pm1.95$& $57.57\pm0.16$& $72.31\pm0.05 $
	    &  $36.64\pm0.05$& $47.26\pm0.13 $& $52.14\pm0.12$ &$66.28\pm0.05$\\
	    ANIL~\cite{ANIL} & $49.61 \pm 0.27$ & $64.90\pm 0.63$& $58.34\pm 0.03$& $72.34\pm 0.01$ &  $36.38\pm 0.16$& $47.16\pm 0.08$ & $54.12\pm0.20$ &$65.18\pm0.14$\\
	    BOIL~\cite{BOIL} & $49.82\pm  0.24$ & $67.60\pm  0.01$& $58.69\pm 0.11$& $75.31\pm  0.02$ &  $38.80\pm 0.03$& $51.42\pm 0.20 $ & $58.03\pm0.16$ &$72.00\pm0.32$\\
	    Reptile~\cite{Reptile} & $47.07\pm0.26$ & $62.74\pm0.37$&$-$&$-$&$-$&$-$&$-$&$-$ \\
	    Prototypical Net~\cite{PrototypicalNetwork}  & $49.42\pm0.78$ & $68.20\pm0.66$&$55.50\pm0.70$&$72.00\pm0.60$&$35.30\pm0.60$&$48.60\pm0.60$& $-$ &$-$ \\
	    Meta-MinibatchProx~\cite{MinibatchProx} & $48.51 \pm 0.92$&$64.15 \pm 0.92$&$-$&$-$&$-$&$-$&$-$&$-$ \\
	    \hline
	    TPN~\cite{TPN}& $54.41\pm0.49$ & $69.54\pm0.33$&$63.53\pm0.28$ &$74.03\pm0.90$&$38.11 \pm0.13$&$48.48\pm0.27$& $-$ &$-$\\
	    Reptile $+$ BN~\cite{Reptile} & $49.97\pm0.32$ & $65.99\pm0.58$&$-$&$-$ &$-$&$-$&$-$&$-$\\
	    Relation Net $+$ BN~\cite{RelationNetwork}  & $50.44\pm0.82$ & $65.32\pm0.70$ &$-$&$-$&$-$&$-$&$-$&$-$\\
	    Meta-MinibatchProx $+$ BN~\cite{MinibatchProx}& $50.77 \pm 0.90$&$67.43 \pm 0.89$&$-$&$-$ &$-$&$-$&$-$&$-$\\
	    MAML $+$ BN~\cite{MAML} & $48.70\pm1.84$ &$63.11\pm0.92$&$-$&$-$&$-$&$-$&$-$&$-$ \\
	    MAML$++$ $+$ BN~\cite{MAML++} &$52.15\pm0.26$ & $68.32\pm0.44$&$-$&$-$ &$-$&$-$&$-$&$-$\\
	    \hline
	    GP-MAML (Ours)  & $52.71\pm0.20$ & $68.06 \pm 0.62$ &$64.03\pm0.06$&$75.60\pm0.27$& $38.57\pm0.36$&$48.50\pm0.52$& $56.28\pm0.24$ &$68.76\pm0.18$\\
	    GP-ANIL (Ours)  & $\mathbf{55.92\pm0.50}$ & $70.73 \pm 0.59$ &$65.66\pm0.25$&$75.08\pm2.57$&$38.95\pm0.03$&$51.16\pm0.19$& $57.54\pm0.06$ &$69.48\pm0.14$\\
	    GP-BOIL (Ours)  & $55.55\pm0.10$ & $\mathbf{71.36 \pm 0.12}$&$\mathbf{66.55\pm0.05}$ &$\mathbf{78.50\pm0.34}$&$\mathbf{41.80\pm0.12}$&$\mathbf{53.17\pm0.04}$& $\mathbf{61.55\pm0.08}$ &$\mathbf{75.18\pm0.26}$\\
	    \hline
    \end{tabular}
    }
    \end{center}
    \label{tab:Performance}
\end{sidewaystable}

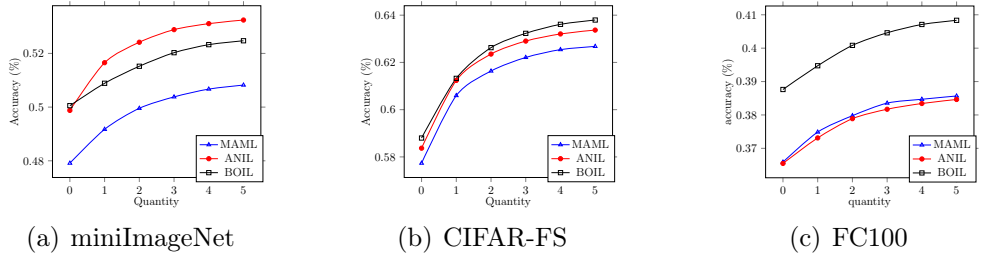
\begin{figure}[htbp]
\centering
\subfigure[miniImageNet]{
\centering
\begin{minipage}[t]{0.3\linewidth}
\centering
\begin{tikzpicture}[scale=0.4] 
\begin{axis}[
    xlabel=Quantity, 
    ylabel=Accuracy (\%), 
    tick align=outside, 
    legend style={at={(0.85,0.245)},anchor=north} 
    ]
\addplot[smooth,mark=triangle,blue] plot coordinates { 
    (0,0.479115)
    (1,0.491727)
    (2,0.499557)
    (3,0.503801)
    (4,0.506711)
    (5,0.508204)
};
\addlegendentry{MAML}
\addplot[smooth,mark=*,red] plot coordinates { 
    (0,0.498767)
    (1,0.516541)
    (2,0.524165)
    (3,0.528852)
    (4,0.531091)
    (5,0.532436)
};
\addlegendentry{ANIL}
\addplot[smooth,mark=square,black] plot coordinates { 
    (0,0.500531)
    (1,0.508877)
    (2,0.515217)
    (3,0.520269)
    (4,0.523263)
    (5,0.524713)
};
\addlegendentry{BOIL}
\end{axis}
\end{tikzpicture}
\end{minipage}%
}%
\subfigure[CIFAR-FS]{
\centering
\begin{minipage}[t]{0.3\linewidth}
\centering
\begin{tikzpicture}[scale=0.4] 
\begin{axis}[
    xlabel=Quantity, 
    ylabel=Accuracy (\%), 
    tick align=outside, 
    legend style={at={(0.85,0.245)},anchor=north} 
    ]
\addplot[smooth,mark=triangle,blue] plot coordinates { 
    (0,0.577347)
    (1,0.605961)
    (2,0.616347)
    (3,0.6221)
    (4,0.62542)
    (5,0.626777)
};
\addlegendentry{MAML}
\addplot[smooth,mark=*,red] plot coordinates { 
    (0,0.583699)
    (1,0.612383)
    (2,0.623535)
    (3,0.629036)
    (4,0.632055)
    (5,0.633689)
};
\addlegendentry{ANIL}
\addplot[smooth,mark=square,black] plot coordinates { 
    (0,0.587991)
    (1,0.613256)
    (2,0.626261)
    (3,0.632312)
    (4,0.636125)
    (5,0.637945)
};
\addlegendentry{BOIL}
\end{axis}
\end{tikzpicture}
\end{minipage}%
} 
\subfigure[FC100]{
\centering
\begin{minipage}[t]{0.3\linewidth}
\centering
\begin{tikzpicture}[scale=0.4] 
\begin{axis}[
    xlabel=quantity, 
    ylabel=accuracy (\%), 
    tick align=outside, 
    legend style={at={(0.85,0.245)},anchor=north} 
    ]
\addplot[smooth,mark=triangle,blue] plot coordinates { 
    (0,0.365883)
    (1,0.374868)
    (2,0.379775)
    (3,0.383541)
    (4,0.384697)
    (5,0.38568)
};
\addlegendentry{MAML}
\addplot[smooth,mark=*,red] plot coordinates { 
    (0,0.365473)
    (1,0.373117)
    (2,0.378883)
    (3,0.381709)
    (4,0.383428)
    (5,0.384633)
};
\addlegendentry{ANIL}
\addplot[smooth,mark=square,black] plot coordinates { 
    (0,0.387624)
    (1,0.394736)
    (2,0.400836)
    (3,0.4046)
    (4,0.407099)
    (5,0.408344)
};
\addlegendentry{BOIL}
\end{axis}
\end{tikzpicture}
\end{minipage}
}%
\centering
\caption{Performance for various number of selected pseudo query data per class. It shows that under the setting of adaptive picking, the model performance increases steadily with the increment on the number of pseudo-label samples.}
\label{fig:adaptivePicking}
\end{figure}

The result shows that the pseudo query data can be well adapted to MAML, ANIL and BOIL after Adaptive Picking, 
We also calculate the accuracy of the model with an increasing number of samples picked for each category under the setting of Adaptive Picking. The results are shown in Fig.~\ref{fig:adaptivePicking}, indicating that more pseudo query data per class is beneficial for performance. 

\subsubsection{Make the pseudo query data more balanced by considering the relationship among samples.}~
The above experiments show two key factors for better performance: the model's feature extractor and the method of labelling and picking pseudo query data.
Based on these observations, we propose GP-MAML, GP-ANIL, and GP-BOIL by generating pseudo-labels for MAML, ANIL, and BOIL, respectively. 

Specifically, when labeling the query set data, we use the body of MAML, ANIL, or BOIL and the graph construction module of TPN to get a weighted graph, and then use label propagation to label data in the query set all at once, which means we can make the query data more balanced by considering the relationship between query set samples. In the end, we use adaptive picking to select pseudo query data which is then used to re-train the model. We apply this framework in both meta-training and meta-testing.
The results shows that our GP-MAML, GP-ANIL and GP-BOIL outperform MAML, ANIL and BOIL, respectively. GP-BOIL outperforms others because the BOIL's body can update its parameters to extract more efficient representations through task-specific adaptation, which is called the \textit{representation change}.
Our model also outperforms TPN because our model can utilize information from the unlabeled query set to re-train the network in a targeted manner in meta-testing, but it is difficult for TPN to perform in this way.
The results of our method are shown in Table \ref{tab:Performance}. 


\section{Conclusion}
\label{05Conclusion}
In this work, we generate pseudo-labels by using label propagation with adaptive picking, introduce transductive methods to typical inductive methods, \ie, the MAML series, and thereby improving their performance. 
We start by taking the pseudo query data into account, addressing the problem that inductive methods can not fully utilize information of the query set. Since the classifier of inductive methods is sensitive to the pseudo query data, we employ feature extractor and label propagation to label the query set. However, as the number of samples in different class of pseudo query data varies substantially, the pseudo query data can not be directly used to re-train the network. To address this issue, we propose a simple yet effective method called adaptive picking to select samples from distinct classes with balanced quantity. Experiments show that when MAML, ANIL, and BOIL are re-trained with pseudo-labeled data, 
they are all boosted with  higher performance, demonstrating the broad applicability of our method.

There are certain observations gained from our work.
First, the query data can be well utilized by pseudo-labeling. 
Second, the imbalances in pseudo query data would have negative impact on models that are very sensitive to data.
Third, the classifier of MAML is not appropriate for pseudo-labeling, and other ways based on the feature extractor of MAML should be considered.
We will follow this line for further exploration in future works. 





\bibliographystyle{splncs}
\bibliography{main.bbl}

\end{document}